\documentclass[10pt, a4paper]{article}

\usepackage{lrec-coling2024} %

\usepackage[normalem]{ulem}
\usepackage{enumitem}
\usepackage{graphicx}
\usepackage{amsmath}
\usepackage{amssymb}
\usepackage{algorithm}
\usepackage[noend]{algpseudocode}
\usepackage{caption}
\usepackage{subcaption}
\usepackage{float}
\usepackage{colortbl}
\usepackage{xcolor}
\usepackage{spverbatim}

\title{Sharing the Cost of Success: A Game for Evaluating and Learning Collaborative Multi-Agent Instruction Giving and Following Policies}

\name{    %
    Philipp Sadler{\rm ,} 
    Sherzod Hakimov {\rm and} 
    David Schlangen
}

\address{    %
    CoLabPotsdam / Computational Linguistics\\ 
    Department of Linguistics, University of Potsdam, Germany\\
    firstname.lastname@uni-potsdam.de\\
    }

\abstract{
In collaborative goal-oriented settings, the participants are not only interested in achieving a successful outcome, but do also implicitly negotiate the effort they put into the interaction (by adapting to each other). In this work, we propose a challenging interactive reference game that requires two players to coordinate on vision and language observations. The learning signal in this game is a score (given after playing) that takes into account the achieved goal and the players' assumed efforts during the interaction. We show that a standard Proximal Policy Optimization (PPO) setup achieves a high success rate when bootstrapped with heuristic partner behaviors that implement insights from the analysis of human-human interactions. And we find that a pairing of neural partners indeed reduces the measured joint effort when playing together repeatedly. However, we observe that in comparison to a reasonable heuristic pairing there is still room for improvement---which invites  further research in the direction of cost-sharing in collaborative interactions.
 \\ \newline  }

\begin{document}

\maketitleabstract

\section{Introduction}

Recent advances in natural language processing have led to language model-based systems that, at least at first sight, seem to do a good job at creating natural dialogue behaviour.
However, the conversations with these models are often still very verbose (lengthy responses) and brittle (necessity to wait for response completion). In contrast, \citet{clark_referring_1986} observed that humans in a collaborative situation used the language as a coordination device (joint action; \citet{clark_using_1996}) and that an adaption process takes place which is driven by effort reduction. In their experiments, a director instructed a listener to put cards with figures on them in a specific order without seeing the listener's cards. The observation was that the average number of speaking turns taken by the director per figure drastically reduced over the trials while the success outcomes stayed high. Thus, the later trials did not just lead to the desired outcome but were also more \textit{efficient}.

\begin{figure}[t]
    \begin{center}
        \includegraphics[width=0.43\textwidth]{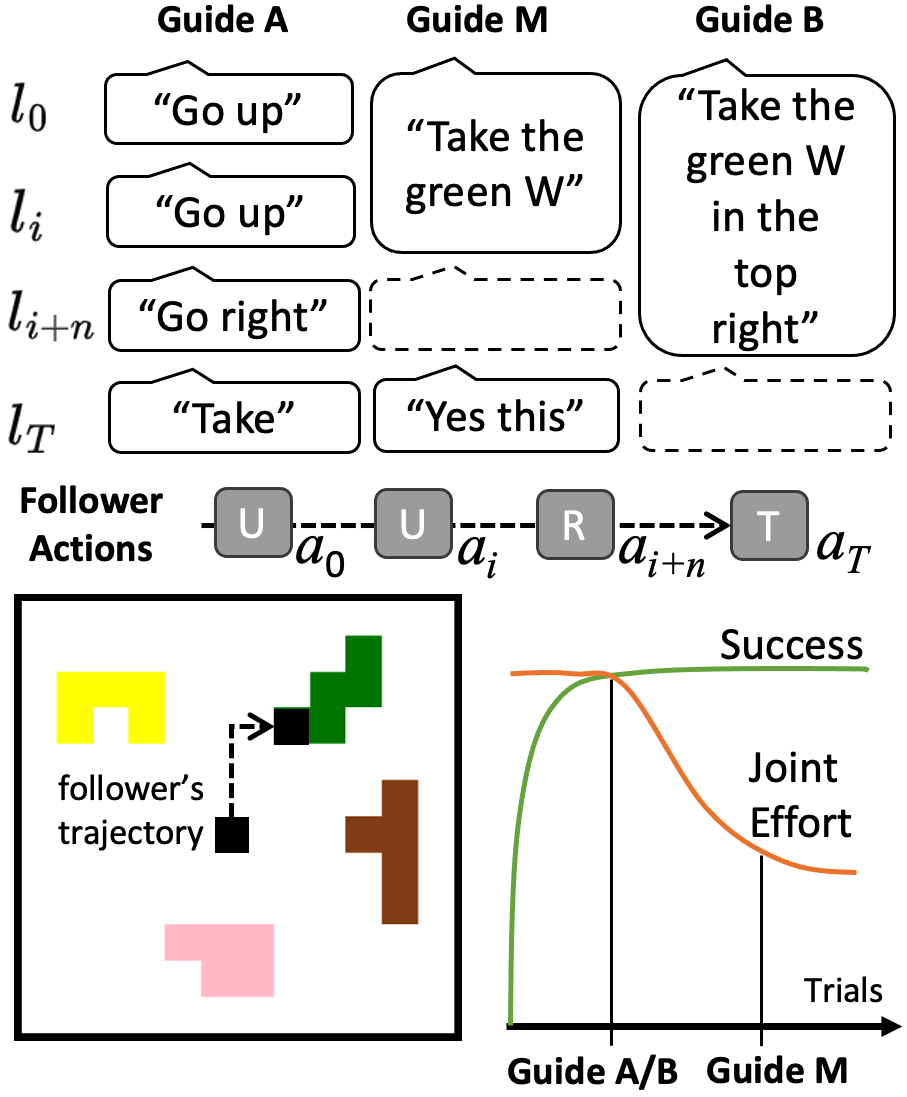}
        \vspace{-0.3cm}
    \end{center}
    \caption{A guide and a follower observe the board with the pieces and the follower's gripper (the black dot). An optimal trajectory of actions for the follower would be: up (U), up, right (R), and take (T). The best strategy for the guide lies assumably in the middle (M) of the extremes (A/B) where the guide refers to a piece initially with $l_0$ and stays silent at until confirming the follower's choice with $l_T$. This strategy shares the cost for success between both. %
    }
    \vspace{-0.5cm}
    \label{fig:example_board}
\end{figure}

Now, imagine a situation as sketched in Figure~\ref{fig:example_board}. An instruction giver guides a follower towards a piece that must be taken on a virtual board, but there are various other pieces which might distract the follower. A strategy for the guide could be to use short phrases and perform remote control (Guide A). The main effort stays with the guide, which has to provide accurate navigation instructions while the follower executes them without incurring any own planning effort costs. Another extreme would be a strategy (Guide B) where the guide initiates the interaction with a very detailed instruction and then stays silent. This puts most of the cognitive load on the follower's side which might now hesitate or actually take the wrong piece after all. The best strategy presumably lies in the middle (Guide M) where the guide -- after having seen different boards previously and having interacted with the follower multiple times -- initiates the interaction with a longer phrase but provides useful feedback when necessary. While all these strategies can be successful, the latter is the one that shares the cost of success the best between the partners. Such capabilities would be essential for future assisting agents to take a helpful part in society someday.%

However, we notice that current research on language and vision coordination problems seems to neglect (a) the notion of required effort for the production and delivery of instructions and (b) the incremental aspects of the interaction. In vision and language navigation \cite{chevalier-boisvert_babyai_2019,DBLP:conf/emnlp/NguyenD19,DBLP:conf/nips/FriedHCRAMBSKD18} a follower receives at each time step a (possibly lengthy) instruction that contains all relevant information. In interactive sub-goal generation, a planning model comes up with a goal formulation in ``no-time'' (a single step) \cite{chane-sane_goal-conditioned_2021,sun_adaplanner_2023,lee_controllable_2023}. And in multi-agent environments, agents typically coordinate without using natural language at all \cite{bard_hanabi_2020,samvelyan_starcraft_2019,pan_mate_nodate}.

Can neural agents agree on a cooperative strategy that shares the cost of success more equally? %
In this work, we put the incremental aspect of the language and vision coordination problem to the fore again and weight an agent's actions by its assumed effort and time costs.
Thus, agents have to trade off the production of costly but informative actions with the overall outcome of the game.
To study neural agent capabilities under this constraint
\begin{itemize}
    \itemsep0cm
    \item we propose a challenging reference game where two players have to coordinate on the selection of a piece among various distractors while the actual target piece is only known to one of them (the guide) and only the other can perform the selection (the follower), 
    \item establish a strong baseline performance with heuristic partners that implement insights from the analysis of human-human interactions
    \item and show that neural partners in a multi-agent setting indeed strive towards an presumably more human-like strategy when effort matters.
\end{itemize}

\section{A Game for Evaluating and Learning Collaborative Multi-Agent Policies}
\label{sec:game}

\begin{figure}[t]
    \begin{center}
        \includegraphics[width=0.40\textwidth]{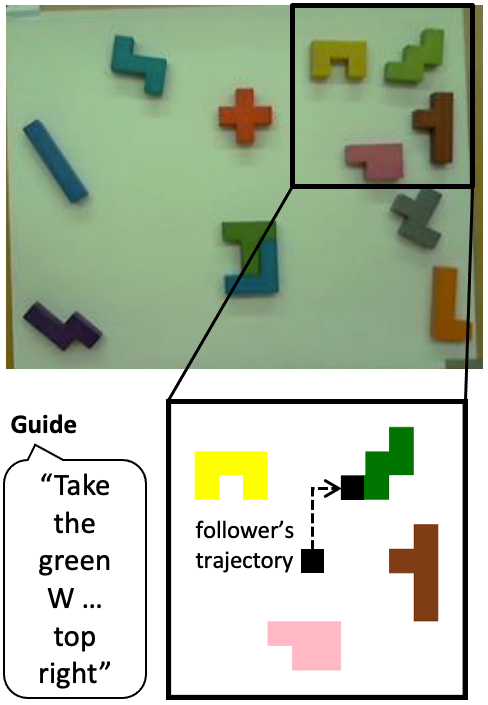}
        \vspace{-0.3cm}
    \end{center}
    \caption{An example from \citet{zarries_pentoref_2016} who found that a reference game leads to diverse language production on the guide's side. To study the aspects of cost sharing in such a collaborative interaction with neural agents, we propose \textsc{CoGRIP} along with a generator for virtual boards that eases the application of data-driven learning methods.}
    \vspace{-0.5cm}
    \label{fig:example_real}
\end{figure}

We propose a \textbf{Co}llaborative \textbf{G}ame of \textbf{R}eferential and \textbf{I}nteractive language with \textbf{P}entomino pieces (\textsc{CoGRIP}) to evaluate and learn neural policies for the aspect of cost sharing a multi-agent setting. In \textsc{CoGRIP} two players are forced to work together because of their asymmetry in knowledge and skill. A guide uses language utterances to instruct a follower to select a puzzle piece (Pentomino; \citet{golomb_1996}). The guide can provide utterances but cannot move the gripper. The follower can move the gripper but is not allowed to provide an utterance.  \citet{zarries_pentoref_2016} found that such a setting leads to diverse language production on the guide's side. For example, there are references with delayed positional descriptions like ``then you take the green W ... top right``, detailed references like ``the green object that looks like a T top left in the corner``, directional reinforcements like ``more to the left'' and confirmations like ``exactly''. We virtualize this setting as shown in Figure~\ref{fig:example_real} for better control and to apply neural learning algorithms in a multi-agent setting. And we frame this as a game where both players receive a score after playing that represents their success and the effort spent for completion.
Next, we explain the details of the game, its scoring and the prepared instances.

\paragraph{Actions.} Formally, the guide's action space $\mathcal{A_\text{G}}$ spans all possible utterances of length $L$ that are possible given the vocabulary $V$ (in English) and includes an action for \texttt{silence}. Likewise the follower's action space $\mathcal{A_\text{F}}$ contains an action for hesitation (\texttt{wait}) and actions for movements (\texttt{left, right, up, down}) as well as an action to \texttt{take}. A board is internally represented as a grid of $M \times M$ tiles and the gripper can only move one tile at a time step. The gripper can move over pieces, but is not allowed to leave the boundaries of the board.

\paragraph{Effort.} The efforts of the players represent the costs for success. We approximate the efforts based on the empirical observations from \citet{zarries_pentoref_2016}. The transcripts of the experiments suggest to group the guide's utterances into five categories to which we attach an effort estimate to each of the categories as follows:

\vspace{-.1cm}
\begin{equation}
 E_{\text{G}} = \sum_{t=1}^T\begin{cases}
 0, & \text{if $a_t$ $\in$ \{\texttt{silence}\}} \\ 
 1.0, & \text{if $a_t$ $\in$ \{\texttt{confirm,decline}\}}\\
 2.0, & \text{if $a_t$ $\in$ \{\texttt{directive}\}}\\ 
 3.0, & \text{if $a_t$ $\in$ \{\texttt{reference}\}}
 \end{cases}
\end{equation}

so that $E_{\text{G}}$ represents the guide's effort in an episode with $T$ steps. These action-based costs follow the assumed cognitive load for producing the according utterances: Here, staying \texttt{silent} (only watching) is the zero baseline. A \texttt{confirm} (``yes this way'') carries the similar meaning (of detecting an action towards the goal) with the additional function of re-assuring the follower to act correctly by the cost of producing a short phrase. A \texttt{decline} phrase (``not this piece'') signals the contradiction between the guides predicted action and the follower ones with the prospect of a follower's correction or if not, it buys time to produce a more demanding instruction. Such an instruction could be a \texttt{directive} (``go left/right/up/down'', ``take'') which requires the comparison between the gripper's position and the target piece. Or even comparing all pieces with each other to produce a \texttt{reference} (``take the green W'') which is reflected with the highest effort cost. For the follower we approximate the effort $E_{\text{F}}$ with

\vspace{-.1cm}
\begin{equation}
 E_{\text{F}} = \sum_{t=1}^T\begin{cases}
 0, & \text{if $a_t$ $\in$ \{\texttt{wait}\}} \\ 
 2.0, & \text{if $a_t$ $\in$ \{\texttt{movement}\}}\\
 3.0, & \text{if $a_t$ $\in$ \{\texttt{take}\}}\\ 
 \end{cases}
\end{equation}

based on the assumed energy costs for performing the action (physically), i.e.,  lifting an object is harder than moving on a plane.

\begin{figure*}[t]
    \begin{center}
        \includegraphics[width=0.99\textwidth]{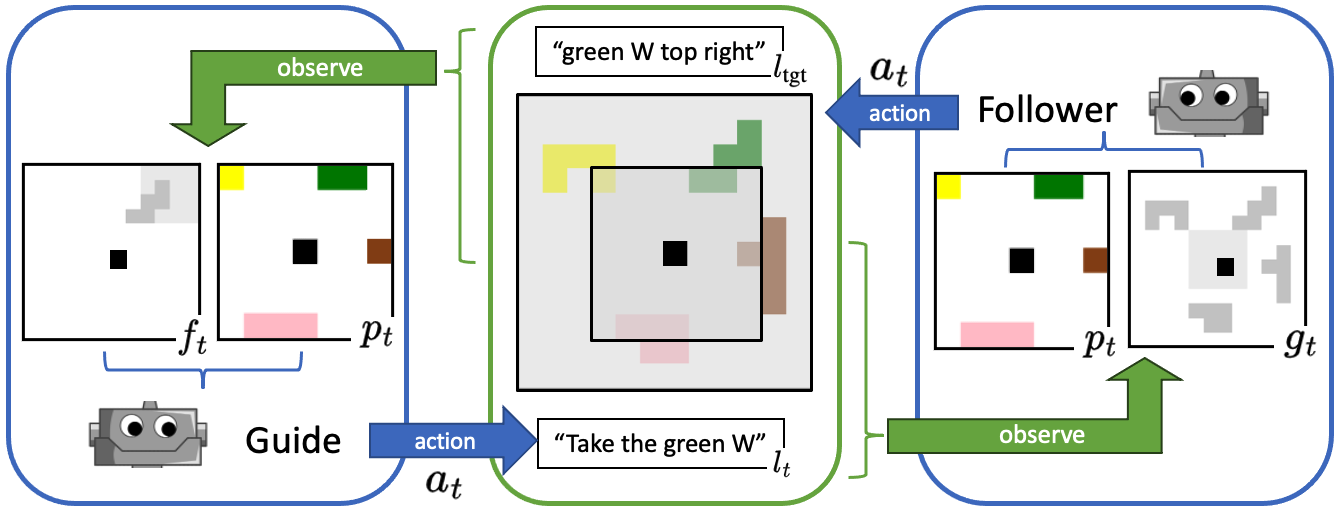}
        \vspace{-0.3cm}
    \end{center}
    \caption{The general information and decision-making flow during an episode of the reference game. The guide observes a constant textual target piece descriptor $l_\text{tgt}$, the partial view $p_t$ and a peripheral overview $g_t$ of the scene. Given this, the guide chooses to produce a language action $a_t$ which could mean ``silence'', a word, a phrase or a sentence that gets translated into an utterance $l_t$.
    The follower receives the utterance $l_t$, the partial view $p_t$ and a peripheral overview $f_t$. Given this, the follower performs an action $a_t$ that results into waiting, a movement (which changes the visual state) or an attempt to take a piece. The game ends when any piece is taken or the maximal number of time-steps $T_{\text{max}}$ is reached.
    }
    \label{fig:gameflow}
   \vspace{-0.3cm}
\end{figure*}
\paragraph{Score.} We measure the quality of an episode of the game with a scoring function that follows the reward formulation of \citet{chevalier-boisvert_babyai_2019}

\vspace{-.1cm}
\begin{equation}
 \text{S}(x) = 1 - 0.9 * (x / T_{\text{max}})   
\end{equation}

where $T_{\text{max}}$ is a hyper-parameter that determines the maximal number of possible time steps in an episode of the game. Now, the quality score of an episode is given by the required time steps $T$ to reach a terminal state $S_{\text{Time}}=S(T)$, the joint effort score $S_{\text{Effort}}$ and the overall outcome of the game $S_{\text{Outcome}}$ which is $+1$ when the correct piece or a penalty of $-1$ if the wrong or no piece has been taken at all, so that:
\vspace{-.1cm}
\begin{equation}
 \text{S}_{\text{Game}} = (\text{S}_{\text{Time}} + \text{S}_{\text{Effort}}) / 2 + \text{S}_{\text{Outcome}} 
\end{equation}

where $\text{S}_{\text{Effort}} = (\text{S}(E_{\text{G}}) + \text{S}(E_{\text{F}})) / 2$. Given this formulation, the players have to to be active (not just wait until $T_{\text{max}}$ is reached), achieve the goal (receive $S_{\text{Outcome}}=+1$) and reduce their individual efforts (stay mostly \texttt{silent} or \texttt{wait} when the utterance is not understood) to reach the highest score. Thus the score ranges from about $-2$ (high effort, long and failure) to $+2$ (low effort, quick and successful).

\paragraph{Game Instance.} %
An instance of the reference game is defined by the size of the board, a target piece and numerous distractor pieces. The appearance and position of the pieces is derived from symbolic piece representations: a tuple of shape ($7$), color ($6$), and position area ($9$). We use $315$ ($7 \cdot 6 \cdot 9$ minus a holdout) of these symbolic pieces to create game instances and split them into distinct sets, so that the target pieces for the testing tasks differ from the ones seen during training (they might share color and shape but are, for example, positioned elsewhere). We provide $1750$ training, $210$ validation, and $245$ testing instances of the task for three board sizes ($12$, $21$, $27$). On these boards, a piece occupies five adjacent tiles and is not allowed to overlap with another one.

\paragraph{Evaluation.} We quantify the performance on the task for a particular pair of follower and guide by letting them play all test game instances (where the follower always starts in the center of a map). We compute the achieved scores for these $N$ testing episodes and average them to constitute the mean task score (mTS) for a pair of guide and follower. Furthermore, we are interested in the mean success rate (mSR) as the number of episodes where the correct piece was selected

\vspace{-0.4cm}
\begin{equation}
\text{mSR}=\frac{1}{N} \sum^{N}_{i=1}{s_i} \text{ where } s_i=\begin{cases}
    1, & \text{for correct piece} \\
    0, & \text{otherwise}
\end{cases}
\end{equation}

as well as the mean episode length (mEPL) as the number of time steps needed to take a piece (with the upper bound $T_{\text{max}}$) and the mean joint effort spent by the pair at each time step (mJE)

\begin{equation}
\text{mJE}=\frac{1}{N} \sum^{N}_{i=1}{\frac{(E_{\text{G}_i} + E_{\text{F}_i})/2}{T_i}} 
\end{equation}

which ranges from $0$ to $3$ (from the guide is always silent and the follower always waits to the guide utters a reference and follower performs take at each time step in an episode).

\section{Learning Neural Policies for Sharing the Cost of Success}

Along with the newly proposed game for cost sharing, we determine baseline performances for neural policies and heuristic ones (which bootstrap them). The neural policies are supposed to learn the means of success in the game solely by playing with the partner. Here, the heuristic policies are supposed to help them to learn successful behavior in the reference game. We hypothesize that once the neural policies have learned how to achieve a successful outcome in the game (over a period of many cooperative interactions), a joint effort reduction takes place to achieve an even better score.

\subsection{Problem Formulation}

For our study, we methodologically frame this game as a reinforcement learning problem \cite{Sutton1998} with sparse rewards. Thus, we treat the guide and follower from here on as \textit{agents} that act in an \textit{environment} (the game), which exposes observations to them. At each time-step $t$, given an observation $o_t \in \mathcal{O}$, the guide has to choose an action $a_t \in \mathcal{A_\text{G}}$ such that the overall resulting sequence of actions $(a_0,...,a_t,...,a_T)$ maximizes the sparse reward $\mathcal{R}(o_T)=S_{\text{Game}}$. Similarly, the follower has to choose an action $a_t \in \mathcal{A_\text{F}}$ at each time step to maximize the shared sparse reward. 
The follower and guide agents act at the same time-step but in consecutive order as depicted in Figure~\ref{fig:gameflow}. %
An episode ends when a piece is selected by the follower or $t$ reaches $T_{max}$ so that an episode does not last forever and the trajectories do not become infinitely long.

\begin{table}[b]
\centering
\small
\begin{tabular}{|c|c|c|rr|}
\hline
\textbf{Size} & \textbf{$N_{\text{Pieces}}$} & \textbf{$T_{\text{max}}$} & \multicolumn{1}{c}{\textbf{\# DTA=0}} & \multicolumn{1}{c|}{\textbf{\# DTA$\geq$1}} \\ \hline
\rowcolor[HTML]{EFEFEF} 
12            & 4                  & 30              & 430 / 58                              & 1320 / 187                             \\
21            & 4--8               & 60              & 396 / 58                              & 1354 / 187                             \\
\rowcolor[HTML]{EFEFEF} 
27            & 4--16              & 80              & 360 / 48                              & 1390 / 197                             \\ \hline
\end{tabular}
\caption{The possible number of pieces ($N_{\text{Pieces}}$) for game instances with boards of the respective sizes and the maximal number of time steps ($T_{\text{max}}$). Game instances with board size 12 have always 4 pieces. For the other we choose uniform random from the range of piece amounts. In the majority of training/testing instances, there is at least one distractor (\# DTA$\geq$1) in the same positional area as the target piece e.g.\ both are in the ``top right''.}
\label{table:maps}
\end{table}

\subsection{Observations}

Our intuition is that the overall task can be decomposed into two sub-tasks: First, the agents should agree on the area where the target piece is supposed to be located, e.g., \ the ``top right''. Then, after reaching this area with the follower's gripper, the agents have to coordinate to select the correct piece in that area (as shown in Table~\ref{table:maps},  as there is more than one candidate in the majority of cases). 

Given these assumptions, we provide the learning agents with two visual perceptions of the scene: a lower-resolution peripheral overview ($f_t$ or $g_t$) to agree on the target area by using positional utterances. And a colored higher resolution focus area (a partial view $p_t$, which is also commonly used in other vision-based reinforcement learning problems; \citep{DBLP:journals/tciaig/HuWSTTYL23,DBLP:journals/corr/abs-2306-13831}) to coordinate about the target piece with use of its shape and color attributes. 
The players share the partial view $p_t$ of the scene, which is centered around the gripper location. This can be interpreted as a behavior where the guide focuses on the follower's ``hand''. The overview observations $f_t$ and $g_t$ look slightly different for each agent to account for the knowledge asymmetry. The guide's overview $g_t$ contains a mask of the target piece, the gripper position, and the ground-truth target area in respective channels. At the same time, the follower's overview $f_t$ channels contain a mask for all pieces, the gripper position, and the current area, respectively.
Furthermore, the guide receives at each time step a constant textual description $l_{tgt}$ of the target piece (e.g., ``blue T top right'') while the follower receives the current utterance $l_{t}$ produced by the guide (which could be silence).

\subsection{Model Architecture}

\begin{figure}[t]
    \begin{center}
        \includegraphics[width=0.45\textwidth]{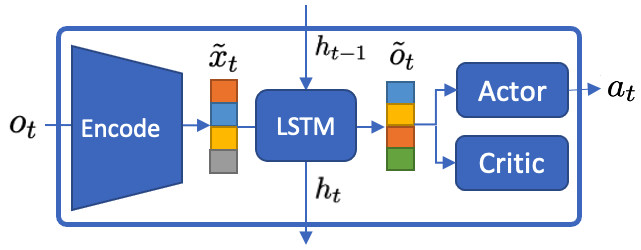}
        \vspace{-0.3cm}
    \end{center}
    \caption{The neural agent's recurrent model architecture includes a memory mechanism (LSTM). At each time-step the observation $o_t$ is encoded and then the resulting embedding $\tilde{x}_t$ is combined with a state representation $h_{t-1}$ of previous time-steps.}
    \label{fig:architecture}
    \vspace{-0.3cm}
\end{figure}

For both guide and follower, we use the same policy architecture as depicted in Figure~\ref{fig:architecture}. While the architecture is the same, the agents receive slightly different observations based on their role in the game (as described above). 
The observations $o_t$ are first encoded into a 128-dimensional feature vector $\tilde{x}_t \in \mathbb{R}$.
Then, the feature vector $\tilde{x}_t$ is fed through an LSTM \citep{hochreiter_long_1997} to produce the memory-conditioned feature vector $\tilde{o}_t$. The LSTM passes a modifiable state vector $h_t$ forward in time (which works as a memory). With this mechanism the follower could memorize already observed utterances (which allows the guide to stay silent), and the guide can anticipate a direction in which the follower is moving (and thus avoid repetitive utterance productions). 

\subsection{Learning Algorithm}

We use \textit{Proximal Policy Optimization} (PPO) \citep{DBLP:journals/corr/SchulmanWDRK17} to learn a parameterized actor-critic policy $\pi(\tilde{o}_t;\theta) \sim a_t$ where the actor predicts a distribution over the action space and the critic estimates the value of the  states. The algorithm basically maximizes the surrogate objective

\vspace{-0.3cm}
\begin{equation}
L(\theta)=\hat{\mathbb{E}}\left[ \frac{ \pi_{\theta}(a_t|s_t) } { \pi_{\theta_{old}}(a_t|s_t) } \hat{A}_t \right]= \hat{\mathbb{E}}\left[ r_t(\theta) \hat{A}_t \right]
\end{equation}

but clips the ratio $r_t(\theta)$ if necessary to stabilize learning. This means when the critic favors the new state $\hat{A}_t>0$ then the policy update is proportional to the ratio $r_t(\theta) \in [0, 1+\epsilon]$ which prevents a too large divergence. And for negative advantages $\hat{A}_t<0$ the probability distribution over action is updated in the opposite direction proportional to the ratio $r_t(\theta) \in [1-\epsilon, \infty]$ which effectively reverts the increase in taking the less favorable action.

We use the recurrent PPO implementation of \textit{StableBaselines3-Contrib} v2.1.0 \citep{stable-baselines3}, because we have the state vector $h_t$ that is passed forward in time as a memory mechanism. The implementation performs back-propagation through time until the first step in an episode.

\subsection{Neural and Heuristic Policies}

Learning cooperative neural agents in this environment from scratch requires a lot from them: the agents must learn (a) that the goal is to take a specific piece and none of the others, (b) the quality score is higher for strategies with less effort, and (c) the visual grounding of utterances themselves (reinforcement signals, references or directives). If training both agents at the same time from scratch, they may solve the task by learning a policy that amounts to a 
language that is inaccessible to humans (emergent languages; \citep{mul_mastering_2019, lowe_pitfalls_2019}) because the vocabulary items can be freely associated with actions (meanings) that are different from what humans understand e.g.\ ``left'' may become (the action) \texttt{right}. Thus, we pair the learning neural agents with fixed heuristic ones that represent a proxy for competent speaker behavior.

\subsubsection{A Neural Follower (NIF)} 

The role of the neural follower (NIF) is to take the piece described by the guide. For this the follower can perform a move, wait, or take action. Formally described, the follower receives at each time-step a vision and language observation $o_t=\{l_t,p_t,f_t\}$ and has to choose an action $a_t \in \{\texttt{wait, left, right, up, down, take}\}$, so that the sequence of actions $(a_0,...,a_t,...,a_T)$ maximizes the sparse reward $\mathcal{R}(o_T)=S_{Game}$. 

\subsubsection{A Heuristic Guide (HIG)}

We pair the neural follower with a heuristic guide behavior (a fixed policy) that has been shown to lead to collaborative success with humans \citep{gotze_interactive_2022}. The heuristic guide always has access to the ground-truth symbolic representations of the pieces on the board and the current gripper position. 
Initially, the guide provides a referring expression $l_0$ that contains the properties necessary to distinguish the target, e.g., ``Take the piece at the top right''. Then the guide provides an utterance $l_i$ at a time-step $t_{>0}$ only when the follower is over a piece or a pre-defined distance/time threshold $R \in \mathbb{N}$ has been exceeded (by comparison of the gripper's last and current position). This can be formally described with the following rules:

\begin{itemize}[leftmargin=*]
    \small
    \itemsep-.1cm
    \item \texttt{wait\_thresh} $\rightarrow$ \texttt{reference} or \texttt{directive(dir)}
    \item \texttt{dist\_thresh} $\rightarrow$ \texttt{dist\_closer} or \texttt{dist\_farther}
    \item \texttt{dist\_closer} $\rightarrow$ \texttt{confirm}
    \item \texttt{dist\_farther} $\rightarrow$ \texttt{decline} or \texttt{directive(dir)}
    \item \texttt{over\_target} $\rightarrow$ \texttt{confirm} or \texttt{directive(take)}
    \item \texttt{over\_other} $\rightarrow$ \texttt{decline} or \texttt{directive(dir)}
\end{itemize}

If none of these rules apply, the guide stays \texttt{silent} $\rightarrow$ \texttt{silence}. Note that the heuristic guide switches between the alternatives on the production side to provide more informative utterances than simply repeating. The production rules follow the effort categorization of Section~\ref{sec:game}. The utterance realization is based on the following templates:

\begin{itemize}[leftmargin=*]
    \small
    \itemsep-.1cm
    \item \texttt{silence} $\rightarrow$ \texttt{<empty string>}
    \item \texttt{confirm} $\rightarrow$ \texttt{Yes this [way|<piece>]}
    \item \texttt{decline} $\rightarrow$ \texttt{Not this [way|<piece>]}
    \item \texttt{directive(\texttt{take})} $\rightarrow$ \texttt{Take <piece>}
    \item \texttt{directive}(\texttt{dir}) $\rightarrow$ \texttt{Go <\texttt{dir}>}
    \item \texttt{reference} $\rightarrow$ \texttt{Take the <IA(PO)>}
\end{itemize}

where \texttt{<piece>} resolves to a piece's color and shape when the current gripper position is located over a piece (or otherwise simply \texttt{piece}). The direction \texttt{<dir>} resolves to the necessary direction of movement. The reference production follows the Incremental Algorithm (\texttt{IA}; a cognitive algorithm by \citet{dale_computational_1995}) that receives a preference over target piece properties (\texttt{PO}).

Here, the heuristic guide is supposed to mimic the intrinsic preference of humans \citep{van_deemter_computational_2016}. The most preferred property is usually the \textit{type} of an object \citep{rosch_cognition_1978}, but in our visual domain all objects are ``pieces'' which makes this attribute uninformative. Although the \textit{shape} could be a proxy for the type, the players would first need to agree on the idea that the pieces represent characters (``W'', ``T'' etc.) and to use it successfully \citep{goudbeek_alignment_2012}. Instead the \textit{color} is likely to be preferred by humans \citep{pechmann_incremental_1989}. Thus, when the follower's gripper is within the target piece area -- meaning that the target piece is most likely visible -- then the heuristic guide prefers color and shape to discriminate the target from its distractors. And otherwise, the guide prefers to mention the target piece's \textit{position} to lead the follower into the target's position.

\subsubsection{A Neural Guide (NIG)}

The neural guide has to produce utterances that help the follower to select the target piece. More formally, the guide receives at each time step an observation $o_t=\{l_{tgt},p_t,g_t\}$ and has to choose an action $a_t \in $ \{\texttt{silence, confirm, decline, left, right, up, down, take, pcs, psc, cps, csp, spc, scp}\} such that the overall resulting sequence of actions $(a_0,...,a_t,...,a_T)$ maximizes the sparse reward $\mathcal{R}(o_T)=S_{Game}$. The chosen actions are realized as utterances with the same mechanism that is used for the heuristic guide to reduce the burden on action space exploration. Note that the actions can be grouped into the five effort categories from Section~\ref{sec:game} where \texttt{directive}'s are \texttt{left, right, up, down, take} and \texttt{reference}'s are the preferences orders \texttt{pcs, psc, cps, csp, spc, scp} (\texttt{c}=color, \texttt{s}=shape, \texttt{p}=position). 

\begin{table*}[t]
\centering
\small
\begin{tabular}{|c|rrrr||rrrr|rrrr|}
\hline
\textbf{}        & \multicolumn{4}{c||}{\textbf{12x12}}                                                                                                         & \multicolumn{4}{c|}{\textbf{21x21}}                                                                                                         & \multicolumn{4}{c|}{\textbf{27x27}}                                                                                                         \\ \hline
\textbf{Pairing} & \multicolumn{1}{c}{\textbf{mSR}} & \multicolumn{1}{c}{\textbf{mEPL}} & \multicolumn{1}{c}{\textbf{mTS}} & \multicolumn{1}{c||}{\textbf{mJE}} & \multicolumn{1}{c}{\textbf{mSR}} & \multicolumn{1}{c}{\textbf{mEPL}} & \multicolumn{1}{c}{\textbf{mTS}} & \multicolumn{1}{c|}{\textbf{mJE}} & \multicolumn{1}{c}{\textbf{mSR}} & \multicolumn{1}{c}{\textbf{mEPL}} & \multicolumn{1}{c}{\textbf{mTS}} & \multicolumn{1}{c|}{\textbf{mJE}} \\ \hline
\rowcolor[HTML]{EFEFEF} 
HIF-HIG          & \textbf{1.00}                    & 7.16                              & 1.75                             & \textbf{1.36}                     & \textbf{0.99}                    & 13.40                             & \textbf{1.74}                    & \textbf{1.33}                     & \textbf{0.98}                    & \textbf{17.64}                    & \textbf{1.73}                    & \textbf{1.33}                     \\
R=1              & 1.00                             & 6.66                              & 1.76                             & 1.46                              & 1.00                             & 13.02                             & 1.76                             & 1.46                              & 1.00                             & 17.66                             & 1.76                             & 1.46                              \\
R=4              & 1.00                             & 7.66                              & 1.74                             & 1.26                              & 0.97                             & 13.78                             & 1.72                             & 1.19                              & 0.95                             & 17.62                             & 1.69                             & 1.20                              \\ \hline
\rowcolor[HTML]{EFEFEF} 
NIF-HIG          & 0.50                             & 9.30                              & 0.79                             & 1.46                              & 0.26                             & 26.23                             & 0.10                             & 1.46                              & 0.17                             & 41.24                             & -0.18                            & 1.48                              \\
R=1              & 0.57                             & 10.42                             & 0.82                             & 1.60                              & 0.29                             & 29.12                             & 0.06                             & 1.62                              & 0.21                             & 44.62                             & -0.20                            & 1.64                              \\
R=4              & 0.43                             & 8.18                              & 0.75                             & 1.31                              & 0.22                             & 23.33                             & 0.13                             & 1.29                              & 0.13                             & 37.85                             & -0.16                            & 1.32                              \\
\rowcolor[HTML]{EFEFEF} 
HIF-NIG          & \textbf{1.00}                    & \textbf{5.19}                     & \textbf{1.79}                    & 1.51                              & 0.96                             & \textbf{12.22}                    & 1.66                             & 1.77                              & 0.90                             & 20.29                             & 1.46                             & 1.84                              \\ \hline
NIF-PNIG*        & {\underline{0.99}}                       & {\underline{6.15}}                        & {\underline{1.73}}                       & 1.77                              & {\underline{0.95}}                       & {\underline{16.23}}                       & {\underline{1.54}}                       & 1.90                              & {\underline{0.93}}                       & {\underline{23.30}}                       & {\underline{1.47}}                       & 1.94                              \\
\rowcolor[HTML]{EFEFEF} 
PNIF-PNIG        & 0.96                             & 7.20                              & 1.63                             & 1.71                              & 0.87                             & 19.35                             & 1.33                             & 1.79                              & 0.69                             & 37.27                             & 0.77                             & 1.90                              \\
NIF-NIG          & 0.95                             & 8.04                              & 1.57                             & {\underline{1.63}}                        & 0.73                             & 27.61                             & 0.87                             & {\underline{1.77}}                        & 0.55                             & 47.7                              & 0.34                             & {\underline{1.80}}                        \\ \hline
\end{tabular}
\caption{The performance of the neural and heuristic pairings on the test instances. We measure the mean success rates (mSR$\uparrow$), the mean episode length (mEPL$\downarrow$), the mean task scores (mTS$\uparrow$) and the mean joint efforts (mRJE$\downarrow$). The best values for a board size are in \textbf{bold}. The best neural-neural performance is \underline{underlined}. PNIG* was kept frozen during training. PNIF-PNIG evaluated with last checkpoint.
}
\label{table:results}
\vspace{-0.3cm}
\end{table*}

\subsubsection{A Heuristic Follower (HIF)}

The heuristic follower to be paired up with the neural guide should be similarly constrained as the neural follower (working with a partial view) so that both neural agents can play together after training with the heuristic partners. Thus, we took inspiration from \citet{sun_adaplanner_2023} and implemented a limited horizon planner that keeps track of and repeatedly revises a plan with up to $6$ actions (the number of actions that is necessary to reach the diagonal corner of the partial view). The heuristic follower always has access to the ground-truth symbolic representation in the partial view and the current gripper position.

The actions in the plan are associated with a probability $p(a_i) = \text{max}(\phi^{i}, \mathrm{L})$ of being executed where $\phi \in [0,1]$ is a discount factor and $\mathrm{L} \in [0,1]$  a lower threshold. This introduces a notion of \textit{confidence}: either the planned action is executed, or a wait action occurs (hesitation). Furthermore, this conceptualizes that a follower becomes increasingly unsure about the continuation of the plan without receiving feedback. If an utterance is received, then its category is determined, and accordingly, one of five sub-programs is run to alter or revise the plan:

\begin{itemize}[leftmargin=*]

    \item{\texttt{on\_silence}:} the follower executes, with respect to the confidence, the next action in the plan (if available). If the plan is empty, then the follower performs the \texttt{on\_reference} sub-program.
    
    \item{\texttt{on\_confirm}:} the follower sets the confidence for all actions in the current plan to 1. Then, the next action or \texttt{wait} is performed.
    
    \item{\texttt{on\_decline}:} the follower erases the current plan and performs \texttt{wait}.
    
    \item{\texttt{on\_directive}:} the follower parses the utterances for directions or take. If the directive suggests taking, then the current plan is erased, and the \texttt{take} action is executed, assuming that this is the last action to be performed. Otherwise, the plan is overwritten with actions towards that direction, and the next action is executed.
    
    \item{\texttt{on\_reference}:} the follower updates its internal target descriptor (color, shape, position) based on the current utterance (which might be empty when coming from \texttt{on\_silence}). Afterward, the partial view is scanned for candidate coordinates based on the target descriptor, e.g., green coordinates given a reference to ``Take the green piece at the top right''. If the descriptor contains a position that is not yet reached, then moving towards that position is prioritized. Otherwise, if the positional area is unknown or already reached, then the shortest path to a candidate coordinate is established as the new plan. If the follower is already in the target area but has no information about shape and color, then a randomly chosen piece in the view is approached. In other cases, the follower \texttt{wait}s.
\end{itemize}

\subsection{Experimental Setup}

We pair the neural agents with the heuristic ones (HIF($\phi=0.99$) and HIG($R=\{1,4\}$) to bootstrap learning, and for comparison, we also run an experiment where they learn from scratch. We train each pairing on the $12 \times 12$ boards from the training split with four environments in parallel (batch size) and 10 million time-steps in total (for a multi-agent learning this means that each agent trains for 5 million steps). Thus, each board in the training split is seen at least 190 times. Every 100k steps during training, we evaluate the policies against the validation set. We keep the policies that achieve the highest mean episode reward based on the validation runs for later evaluation on the testing boards. We do this procedure for three different random seeds ($49184$, $92999$, $98506$) and average the results where applicable.

\subsection{Results \& Discussion}

\paragraph{HIF-HIG is a very strong baseline pairing.} The results in Table~\ref{table:results} show that the HIF-HIG pair achieves a 100\% success rate along with the least joint effort ($1.36$) on the $12 \times 12$ test instances and generalizes to bigger map sizes as well. This very strong performance is supposedly achievable \textit{after} a pairing went through an optimization process as observed by \citet{clark_referring_1986} which results in a collaborative strategy where utterances are mutually understood, properly grounded and produced in such a way that the individual effort is reduced without preventing a successful outcome.\footnote{The NIF-HIG pairing does not fulfill this criteria as the low success rate ($50\%$) indicates that the neural follower has not properly learned the goal condition of the game.} The downside of the heuristic policies is that they cannot easily adapt to others or improve further.

\begin{figure}[t]
    \begin{center}
        \includegraphics[width=0.45\textwidth]{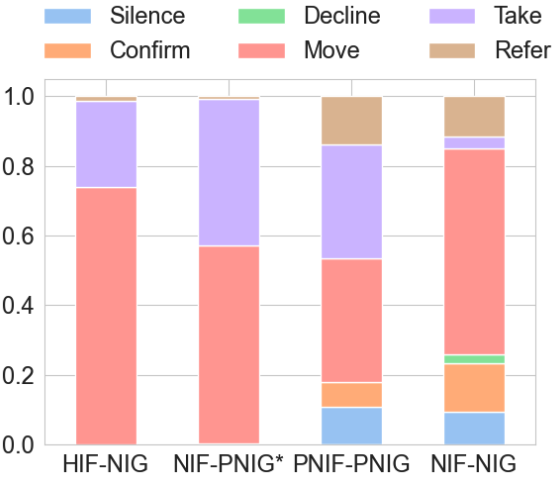}
    \end{center}
    \vspace{-0.3cm}
    \caption{The relative usage of utterance categories per time-step for the guide in various pairings.
    }
    \vspace{-0.1cm}
    \label{fig:speaker_utterances}
\end{figure}

\vspace{-0.3cm}
\paragraph{HIF-NIG pairing exhibits ``Guide A'' strategy.} Thus we pair a learning agent (neural guide) with a heuristic follower that can properly ground the utterances, so that the guide can easier learn to use the intents in a successful way. And indeed the HIF-NIG pair achieves a 100\% success rate on $12 \times 12$ test boards with even less time steps as the heuristic pair ($5.19$) resulting in the highest task score ($1.79$). We also notice that the HIF-NIG pair generalizes to other map sizes. We find that the main reason for this superb performance is that the learnt strategy puts the most effort single-sided onto the guide: it provides a movement directive at almost every time step (see Figure~\ref{fig:speaker_utterances}). As hypothesised in the introduction, although this strategy is highly successfully, it does not result in the least joint effort. And indeed the mean effort is still higher ($1.51$) than the one of the heuristic pair ($1.36$).

\paragraph{NIF-NIG pairings strive towards ``Guide M''.} We hypothesized initially that the best strategy for the guide would be to initially produce a reference and then intervene \textit{only when necessary} (the ``M''iddle way of the extremes of producing an utterance at each time step or only initially). Such a strategy is presumably reached by an \textit{adaption process} between the two collaborators. Since our heuristic policies cannot adapt to their counterparts we train also a pairing of learning agents (neural guides and followers). This NIF-NIG pairing achieves a remarkable success rate ($95\%$) on the $12 \times 12$ boards based on a strategy that involves the whole repertoire of utterances (see Figure~\ref{fig:speaker_utterances}). Notably the guide learns to stay silent in almost $10\%$ of the steps which reduces its effort.

Still, the neural policies might converge to a communication protocol that is inaccessible to humans. Thus, we pair a neural follower (NIF) with the pre-trained neural guide and keep the guide's parameters frozen (PNIG*). This guide learnt to produce the utterances in a way that humans (the heuristic follower) would understand. The results show that the neural follower successfully adapts to the ``Guide A'' strategy: The utterance production is about the same for HIF-NIG and NIF-PNIG* (see Figure~\ref{fig:speaker_utterances}). Consequently, the pair achieves a similar high success rate ($99\%$) and the shortest episodes ($6.15$) among the neural-neural pairings.

\begin{figure}[t]
    \begin{center}
        \includegraphics[width=0.45\textwidth]{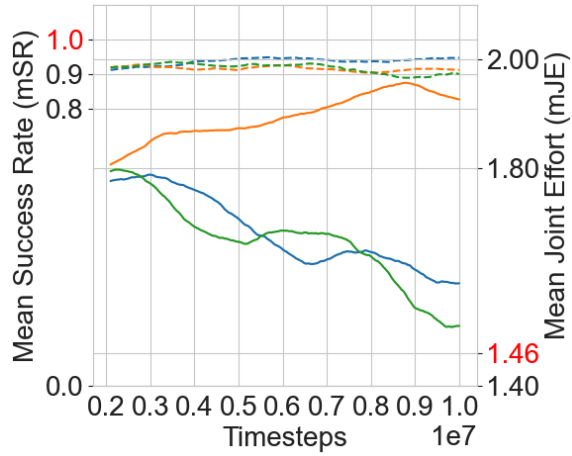}
    \end{center}
    \vspace{-0.3cm}
    \caption{The (smoothed) training curves for the PNIF-PNIG pairing show the mean success rate (dashed lines) stays high and that the mean joint effort (mJE) reduces further (for 2 of 3 seeds). The HIF-HIG performance is indicated by the red ticks.
    }
    \vspace{-0.2cm}
    \label{fig:effort_reduction}
\end{figure}

Now the communication strategy of the neural-neural pairing (PNIF-PNIG) is more accessible to humans, but the joint effort ($1.77$) is above the one from the NIF-NIG ($1.63$). We continue training the pre-trained agents (using their best checkpoints as starting points) in a multi-agent fashion another 10M time-steps, so that the neural agents can further \textit{adapt to each other}. We see that the neural pairings strive towards a strategy that further reduces the joint effort while maintaining the high success rates as shown in  Figure~\ref{fig:effort_reduction}. The best seed achieves a mean joint effort of $1.53$, which is just above the heuristics, and the resulting overall mean efforts ($1.71$) are lower than before ($1.77$). The neural agents that went through the \textit{adaption process} conduct a  strategy that involves more references and also more silence (see Figure~\ref{fig:speaker_utterances}). This indicates that the neural agents  strive towards a ``Guide M'' strategy that shares the cost of success better. %

\begin{table}[b]
\centering
\small
\begin{tabular}{|c|cccc|}
\hline
\textbf{Pairing} & \multicolumn{1}{c}{\textbf{mSR $\uparrow$}} & \multicolumn{1}{c}{\textbf{mEPL $\downarrow$}} & \multicolumn{1}{c}{\textbf{mTS $\uparrow$}} & \multicolumn{1}{c|}{\textbf{mJE $\downarrow$}} \\ \hline
\rowcolor[HTML]{EFEFEF} 
HIF-NIG\textsuperscript{W}        & \textbf{1.00}                             & \textbf{5.71}                              & \textbf{1.80}                             & \textbf{1.13}                              \\
NIF-PNIG\textsuperscript{W}*      & \underline{\textbf{1.00}}                             & \underline{5.93}                              & \underline{1.79}                             & 1.25                              \\
\rowcolor[HTML]{EFEFEF} 
PNIF-PNIG\textsuperscript{W}      & 0.94                             & 7.84                              & 1.61                             & 1.24                              \\
NIF-NIG\textsuperscript{W}        & 0.84                             & 10.48                             & 1.34                             & \underline{1.23}                              \\ \hline
\end{tabular}
\caption{The results for the word-level pairings of guide and heuristic follower on the test boards of size 12. The assumed effort per word is here $1.0$ (and thus not directly comparable with Table~\ref{table:results}).}
\label{table:word_prop_env}
\vspace{-0.5cm}
\end{table}

\paragraph{Interactive language learner mimics ``Guide A''.}

\begin{figure*}[t]
    \begin{center}
        \includegraphics[width=0.90\textwidth]{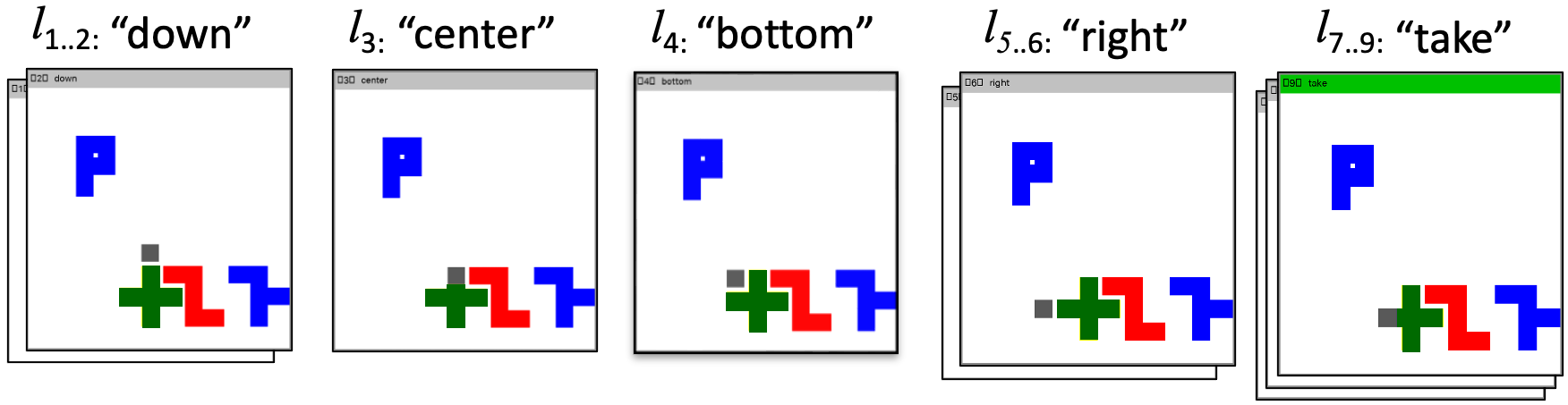}
        \vspace{-0.4cm}
    \end{center}
    \caption{An example episode of the PNIF-PNIG\textsuperscript{W} pair on the validation boards after training. The guide produces a word at each time step to almost ``remote control'' the follower resulting in a high success rate.
    }
    \label{fig:example_episode}
\end{figure*}

The previously described guides produce references by choosing a preference order. This more abstract prediction level allows the guide to focus on learning a useful coordination strategy. Nonetheless, the follower has to understand the actual realization of the reference to perform its actions. Thus we additionally investigate, if a neural guide can learn a useful language production from the interaction alone. We convert ``intent''-actions to words and let a neural guide (NIG\textsuperscript{W}) choose actual property values (colors, shapes and positions) which leads to $24$ ``word''-actions in total. We adjust the heuristic follower to categorize the words correctly and assume an effort of $1$ for a guide's action. A produced word is fed back to the guide in addition to the other observations. The results in Table~\ref{table:word_prop_env} show that the NIG\textsuperscript{W} achieves high success rates. And a qualitative analysis (e.g.\ Figure~\ref{fig:example_episode}) reveals that these results are based on a ``Guide A'' strategy.

\section{Related Work}

As an initial study our work connects ideas from the linguistic interaction field which asks how language shapes an interaction of interlocutors (and vice versa) \citep{gandolfi_mechanisms_2022} and the vision and language field where the actions are visually-grounded \citep{zhang_lovis_2022,dainese_reader_2023}. We realize this connection via the paradigm of reinforcement learning (RL) that introduces a notion of time and allows for incremental processing (which has been recently studied for interactive dictation by \citet{li_toward_2023}). We notice that recent work towards adaptive NLG \citep{ohashi_adaptive_2022} and language feedback \citep{yan_learning_2023} are not visually grounded or approaches that involve vision have no (adaptive) interactive feedback \citep{zhang_lovis_2022, dainese_reader_2023} that has to be learnt from the interaction alone without pre-trained on a pre-collected dataset. Our work is an attempt to connect the research ideas of these fields.\footnote{Source code is publicly available at: \url{https://github.com/clp-research/cost-sharing-reference-game}}

\vspace{-3mm}
\paragraph{Vision and language navigation.}
The proposed reference game shares similarities with vision and language navigation as the follower has to select (navigate to) a specific piece given an utterance. Nevertheless, in navigation tasks, there is usually a lengthy and detailed initial instruction which is accessible at every time step, and the metrics of interest are success rate alone \cite{chevalier-boisvert_babyai_2019}, or additionally episode length \cite{DBLP:conf/emnlp/NguyenD19,DBLP:conf/nips/FriedHCRAMBSKD18}. We are especially interested in the behavior under the constraint of an assumed joint effort and focus on the incremental aspects of language and vision coordination. In our setting, the agents are required to perceptually ground the language to produce a movement (see also \citet{DBLP:journals/corr/abs-2306-13831} as an example of a popular abstract navigation domain) or to produce a language act (here a verbalized intent to reduce space complexity) given the vision inputs at each time-step. 

\vspace{-3mm}
\paragraph{Cooperative multi-agent RL environments.}
Multi-player games present a useful environment to study multi-agent behavior with reinforcement learning, as there are usually well-defined constraints and rewards.
Though to our knowledge, the communication between agents is usually not done via language utterances \cite{bard_hanabi_2020,samvelyan_starcraft_2019,pan_mate_nodate,mohanty_flatland-rl_2020, kurach_google_2020}. The most similar environment we found is from \citet{mordatch2017emergence}, which studied cooperative communication where a listener has to navigate to one of three landmarks. The target is only known by a speaker that can not move. The agents learned from a dense reward signal, which is the distance to the ground-truth landmark. In our game, we only provide a sparse reward and are interested in the behavior after learning to be successful.

\section{Conclusion \& Further Work}

In this work, we proposed a new game to study cooperative multi-agent behavior for cost-sharing, and we presented neural and heuristic policies for learning in this environment. 
We showed that an off-the-shelf learning algorithm (PPO) with a simple reward mechanism (sparse) learns policies that are successful in the game and that continue reducing an assumed joint effort. Nevertheless, the resulting agents lack variety in their coordination strategies (converge to remote control) and still require more effort than a sensible heuristic pairing. Thus our reference game provides a useful foundation and suggests further research in this interesting topic,
so that future neural agents learn more diverse (human-like) language-based coordination behaviors and share the cost of success even better with their interaction counterparts.

\section*{Acknowledgements} We thank the anonymous reviewers for their valuable feedback. This work was funded by the \textit{Deutsche Forschungsgemeinschaft} (DFG, German Research Foundation) – 423217434 (``RECOLAGE'') grant.

\section*{Limitations}

\paragraph{Limits on visual naturalness.} We chose this relatively abstract setting so as to be able to investigate in detail the contribution of each modelling decision. Moving to a more realistic and visually more complex environment is a necessary, but logically later, step. Nevertheless, we think our approach can also be applied to photo-realistic environments \citep{DBLP:conf/nips/RamakrishnanGWM21,DBLP:journals/corr/abs-1712-05474}.

\paragraph{Limits on the visual variety.} The variety of pieces is limited to $7$ different shapes and $6$ different colors. Furthermore, the pieces show no texture but are drawn with a solid color fill. Nevertheless, the visualisations are fast to compute and despite of their simplicity we observed that such a setting produces interesting and complex interactions between a follower and a guide. We leave experimentation on visually even more complex scenes or scenes with ambiguity for future work. 

\nocite{*}
\section{Bibliographical References}\label{sec:reference}

\bibliographystyle{lrec-coling2024-natbib}
\bibliography{lrec-coling2024-paper}

\appendix

\section{Appendix}
\label{sec:appendix}

Robot image in Figure~\ref{fig:example_board} adjusted from \url{https://commons.wikimedia.org/wiki/File:Cartoon_Robot.svg}.
That file was made available under the Creative Commons CC0 1.0 Universal Public Domain Dedication.

\subsection{Observation Details} 

The environment exposes at each time-step all relevant observations so that any combination of the policies can be used within the environment. This means that the neural policy learners use the partial visual observation and the tokenized language utterance along with the positional mask as follows ($|V|=54$ is the vocabulary size, $L=16$ is the maximum sentence length and $M$ is the map size.)

\paragraph{Neural follower observations:}
\begin{itemize}
    \itemsep0em
    \item \texttt{RGB\_PARTIAL} $= \{p_t \in \mathbb{N}_0^{7 \times 7 \times 3} | p_t \leq 255 \}$
    \item \texttt{POS\_FULL\_CURRENT} $= g_t \in \{0,1\}^{M \times M \times 4}$
    \item \texttt{LANGUAGE} $= \{ l_t \in \mathbb{N}^{L} | 0 \leq l_t \leq |V|  \}$
\end{itemize}

where \texttt{RGB\_PARTIAL} is the partial RGB view around the current gripper position, \texttt{POS\_FULL\_CURRENT} contains masks for the board, the current gripper position, the pieces on the board and the current positional area and \texttt{LANGUAGE} contains the last produced utterance.

\paragraph{Neural guide observations:}
\begin{itemize}
    \item \texttt{RGB\_PARTIAL} $= \{p_t \in \mathbb{N}_0^{7 \times 7 \times 3} | p_t \leq 255 \}$
    \item \texttt{POS\_FULL\_TARGET} $= f_t \in \{0,1\}^{M \times M \times 4} $
    \item \texttt{TARGET\_DESC} $= \{ l_{tgt} \in \mathbb{N}_0^{L} | l_{tgt} \leq |V|  \}$
\end{itemize}

where \texttt{RGB\_PARTIAL} is the partial RGB view around the current gripper position, \texttt{POS\_FULL\_CURRENT} contains masks for the board, the current gripper position, the target piece on the board and the target's positional area and \texttt{TARGET\_DESC} contains the tokenized properties of the target.
And the heuristic policies use the symbolic equivalents of the observations as follows:

\paragraph{Heuristic follower observations:}
\begin{itemize}
    \item \texttt{SYM\_PARTIAL} $= \{P_t \in \mathbb{N}_0^{7 \times 7 \times 3} | P_t \leq 255 \}$
    \item \texttt{SYM\_AREA} $= A_t \in \{1,...,9\}$
    \item \texttt{SYM\_POS} $= \{G_t \in \mathbb{N}_0^{2} | G_t \leq M \}$
    \item \texttt{LANGUAGE} $= \{l_t \in \mathbb{N}_0^{L} | l_t \leq |V|  \}$
\end{itemize}

where \texttt{SYM\_PARTIAL} is the partial symbolic view around the current gripper position with the symbolic colors, shapes and object id channels, \texttt{SYM\_AREA} is the symbolic representation of the positional area the gripper is currently in, \texttt{SYM\_POS} are the gripper's current $x,y$-coordinates and  \texttt{LANGUAGE} contains the last produced utterance. 

The symbolic representations for the shapes are: P (2), X (3), T (4), Z (5), W (6), U (7), F(8). The colors are encoded as: red (2), green (3), blue (4), yellow (5), brown (6), purple (7). The 0-symbol is reserved for out-of-world tiles which can occur in the partial view and peripheral view masks. The 1-symbol is reserved for an empty tile. The positional areas are enumerate as: top left (1), top center (2), top right (3), right center (4), bottom right (5), bottom center (6), bottom left (7), left center (8), center (9).

\paragraph{Heuristic guide observations:}
\begin{itemize}
    \item \texttt{SYM\_POS} $= \{G_t \in \mathbb{N}_0^{2} | G_t \leq M \}$
\end{itemize}

where  \texttt{SYM\_POS} are the gripper's current $x,y$-coordinates. In addition, the heuristic guide receives the information about the target's attributes and position at the start of each episode. The distances between two coordinates $(p_1,p_2)$ are calculated as the euclidean distance.

\subsection{Neural Policy Details} 
\label{appendix:agents}

The agents encode two streams of visual inputs: one is the partial visual observation of the scene in colors (RGB) and the other is an overview of the scenes that encodes the position of the gripper on the board and the targeted (or current) positional area. Each of the vision embeddings is input to a FiLM layer that conditions the vision inputs on the target piece descriptor (for the Guide) or the utterance (for the Follower). These language-conditioned vision embeddings are then concatenated and input to the policy network. All model implementations are done in PyTorch v1.13.0 \cite{DBLP:conf/nips/PaszkeGMLBCKLGA19}.

\paragraph{Partial View Encoding.} For the encoding of the partial view we use a CNN with 4 blocks of convolutions, batch norm and relu activations. The first block applies 32 kernels of size 5 with stride 1 and padding. This layer is supposed to learn edges and colors. The second layer applies 64 kernels of size 5 with stride 5 and no padding to shrink the input to the original spatial dimensions of $7\times7$. Then layer 3 and 4 apply 128 kernels of size 3 and padding each resulting in 128 $7\times7$ feature maps that embed the high level visual information of the partial view.

\paragraph{Overview Encoding.} For the encoding of the overview we also use a CNN with 4 blocks of convolutions, batch norm and relu activations. The first block applies 32 kernels of size 1 with stride 1 and no padding. This block is supposed to learn whether the gripper is located in the target area. The other blocks apply 64, 128, 128 kernels of size 3 with padding respectively resulting in 128 $W \times H$ feature maps that embed the high-level positional information of the overview.

\paragraph{Language-conditioning.} The language observations of the agents (the target descriptor for the Guide and the utterances for the Follower) are embedded to 32-dimensional word vectors and then encoded with a GRU which has 128 hidden state dimensions. The last state of the GRU is given as the language encoding to the FiLM layers: one layer that conditions the partial view and one layer that conditions the overview encoding.

\paragraph{Recurrent Policy Network.} The two language-conditioned visual embeddings are added and passed to an LSTM with 128 hidden dimensions. The LSTM embeds the observations over time and can keep track of previous actions. The LSTM's last hidden state is then given to the actor-critic policy network. The actor and the critic are 2-layer feedforward networks where each layer has 64 parameters.

\paragraph{Hyperparameters.}

\begin{table}[h]
    \centering
    \begin{tabular}{| l | r| }
        \hline
         feature\_dims & 128 \\
         normalize\_images & True \\
         shared\_lstm & True \\
         enable\_critic\_lstm & False \\
         n\_lstm\_layers & 1 \\
         lstm\_hidden\_size & 128 \\
         net\_arch & [ [64,64], [64,64] ] \\
        \hline
    \end{tabular}
    \caption{Policy arguments for the the neural agents.}
    \label{tab:agent_hyperparameters}
\end{table}

We use the RecurrentPPO implementation from StableBaselines-Contrib v2.1.0 with the default learning hyper-parameters and the policy parameters as given in Table~\ref{tab:agent_hyperparameters}.

\paragraph{Experiments.}
\label{appendix:experiment}

We trained the pairings in parallel on 8 GeForce GTX 1080 Ti (11GB) where each of them consumed about 4GB of GPU memory. The training of an individual pairing (and seed) for the 5 million steps took about 1 day. 

For the multi-agent training we switched the agent to be updated after each policy update so that the training took 10 million steps in total.

\subsection{Incremental Algorithm (IA)} 
\label{appendix:heuristics}

Both the neural and heuristic guide employ the Incremental Algorithm for referring expression generation via the selection of a preference order (except the NIG\textsuperscript{W} which produces the actual words directly).

\paragraph{Algorithm.} The Algorithm~\ref{alg:ia}, in the formulation of \cite{dale_computational_1995},
is supposed to find the properties that uniquely identify an object among others given a preference over properties. To accomplish this the algorithm is given the property values $\mathcal{P}$ of distractors in $M$ and of a referent $r$. Then the algorithm excludes distractors in several iterations until either $M$ is empty or every property of $r$ has been tested. During the exclusion process the algorithm computes the set of distractors that do \textit{not} share a given property with the referent and stores the property in $\mathcal{D}$. These properties in $\mathcal{D}$ are the ones that distinguish the referent from the others and thus will be returned.

\paragraph{Preference order.} The algorithm has a meta-parameter $\mathcal{O}$, indicating the \textit{preference order}, which determines the order in which the properties of the referent are tested against the distractors. In our domain, for example, when \textit{color} is the most preferred property, the algorithm might return \textsc{blue}, if this property already excludes all distractors. When \textit{shape} is the preferred property and all distractors do \textit{not} share the shape \textsc{T} with the referent, \textsc{T} would be returned. Hence even when the referent and distractor pieces are the same, different preference orders might lead to different expressions.

\begin{algorithm}[h]
\caption{The \textsc{ia} on symbolic properties as based on the formulation by \citet{van_deemter_computational_2016}}\label{alg:ia}
\begin{algorithmic}[1]
\Require{A set of distractors $M$, a set of property values $\mathcal{P}$ of a referent $r$ and a linear preference order $\mathcal{O}$ over the property values $\mathcal{P}$}
\State{$\mathcal{D} \gets \emptyset $}

\For{$P$ in $\mathcal{O}(\mathcal{P})$}
{
\State{$\mathcal{E} \gets \{ m \in M: \neg P(m)$\}}
\If{$\mathcal{E} \ne \emptyset$}
    \State Add $P$ to $\mathcal{D}$
    \State Remove $\mathcal{E}$ from $M$
\EndIf
\EndFor
}
\State{\textbf{return} $\mathcal{D}$}
\end{algorithmic}
\end{algorithm}

\paragraph{Templates.} There are 3 expression templates that are used when only a single property value of the target piece is returned by the Incremental Algorithm (\textsc{ia}): 
\begin{itemize}
    \itemsep0em
    \item \textit{take the [color] piece}
    \item \textit{take the [shape]}
    \item \textit{take the piece at [position]}
\end{itemize}
Then there are 3 expression templates that are selected when two properties are returned:
\begin{itemize}
    \itemsep0em
    \item \textit{take the [color] [shape]}
    \item \textit{take the [color] piece at [position]}
    \item \textit{take the [shape] at [position]}
\end{itemize}
And finally there is one expression templates that lists all property values to identify a target piece:
\begin{itemize}
    \itemsep0em
    \item \textit{take the [color] [shape] at [position]}
\end{itemize}

\subsection{Task Generation}

\paragraph{Symbolic piece splits.} For the task generation we first split the set of the $378$ possible symbolic pieces (a combination of color, shape and position) into different subsets, so that training, validation and testing splits do not overlap. This results into $250$/$30$/$35$ symbolic pieces for training/validation/testing respectively (and a holdout of $63$ symbols that we did not use).

\paragraph{Utterance type-oriented sampling.} Then we iterate through the symbolic pieces in the split and treat each of them as the target piece once. For the target piece we sample a set of distractor pieces in such a way that the IA's reference production would lead to one of the templates (from above) once. This means that per target piece 7 different distractor sets are sampled which leads to $1750$/$210$/$245$ tasks for training/validation/testing respectively. For each task the pieces are put on an initially empty board starting with the target. And then the other pieces are tried to be placed. If a piece cannot be placed on a board without collision, then we choose another coordinate and try this up to 100 times for each placement, until all pieces are placed.

\end{document}